%% file: main.tex
\documentclass[10pt,twocolumn,letterpaper]{article}

\usepackage[pagenumbers]{cvpr} 
\usepackage{overpic}
\usepackage{makecell}
\usepackage{adjustbox} 
\usepackage{multirow}
\usepackage{float}
\usepackage{bbding}
\usepackage{overpic}
\usepackage{times}
\usepackage{epsfig}
\usepackage{graphicx}
\usepackage{amsmath}
\usepackage{amssymb}
\usepackage{booktabs}
\usepackage{xcolor}
\usepackage{enumitem}
\usepackage{lipsum}
\usepackage{diagbox}

\usepackage[ruled,vlined,linesnumbered]{algorithm2e}
\makeatletter
\newcommand{\algorithmfootnote}[2][\footnotesize]{%
  \let\old@algocf@finish\@algocf@finish
  \def\@algocf@finish{\old@algocf@finish
    \leavevmode\rlap{\begin{minipage}{\linewidth}
    #1#2
    \end{minipage}}%
  }%
}
\makeatother

\definecolor{citecolor}{HTML}{0071bc}
\usepackage[pagebackref=false,breaklinks=true,letterpaper=true,colorlinks,citecolor=citecolor,bookmarks=false]{hyperref}

\newcommand{\app}{\raise.17ex\hbox{$\scriptstyle\sim$}}

\definecolor{codegreen}{rgb}{0.0,0.6,0.0}

\definecolor{cvprblue}{rgb}{0.21,0.49,0.74}

\def\paperID{15222} 
\def\confName{CVPR}
\def\confYear{2024}

\title{Panoptic-FlashOcc: An Efficient Baseline to Marry Semantic Occupancy with Panoptic via Instance Center}

\author{
Zichen Yu$^{1}$ \quad Changyong Shu$^{2}$$\textsuperscript{\Envelope}$ \quad Qianpu Sun$^{3}$ \quad Yifan Bian$^{4}$ \\ 
\quad Xiaobao Wei$^{2}$ \quad Jiangyong Yu$^{2}$ \quad Zongdai Liu$^{2}$ 
\quad Dawei Yang$^{2}$ 
\quad Hui Li$^{2}$ 
\quad Yan Chen$^{2}$ \\
$^1$Dalian University of Technology, $^2$Houmo AI, $^3$Tsinghua University, \\
$^4$Harbin Institute of Technology \\
{\tt\small yuzichen@mail.dlut.edu.cn,sqp23@mails.tsinghua.edu.cn,byf1522484664@163.com} \\
{\tt\small \{changyong.shu,xiaobao wei,jiangyong.yu,zongdai.liu,dawei.yang,hui.li,yan.chen\}@houmo.ai}
\vspace{-15pt}
}

\begin{document}

\makeatletter
\vspace{-20pt}
\let\@oldmaketitle\@maketitle
\renewcommand{\@maketitle}{\@oldmaketitle
\centering
    \begin{overpic}[width=1.0\textwidth]{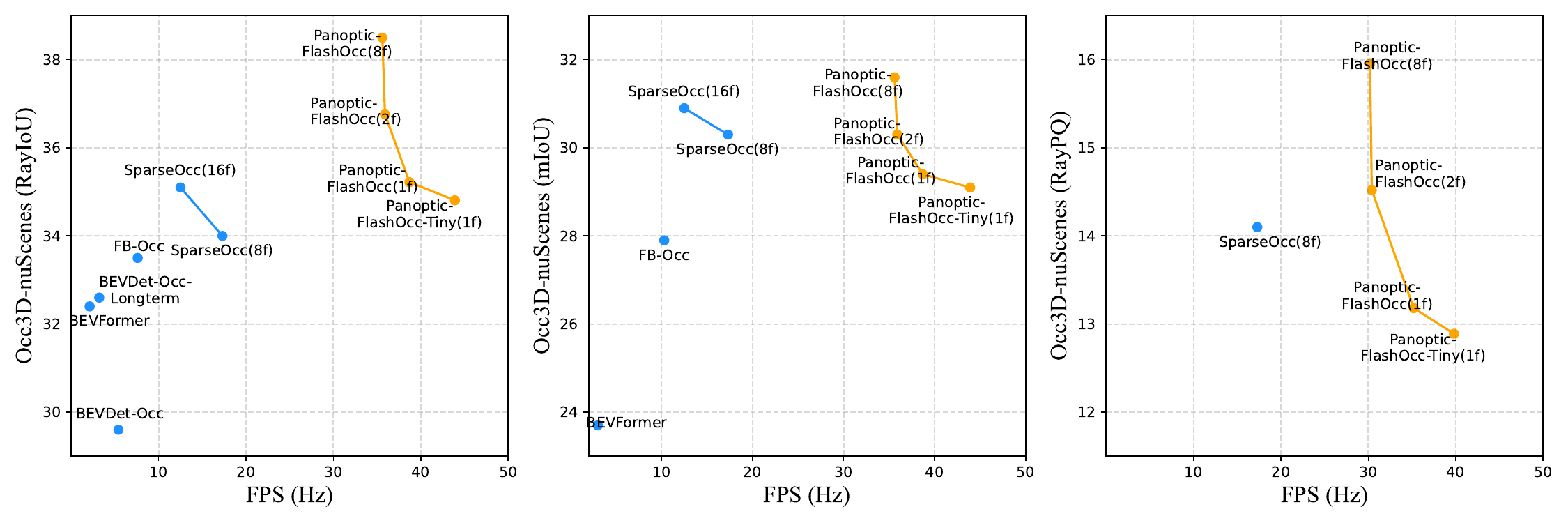}
    \end{overpic}
    \captionof{figure}{
    Trade-off between Speed and Accuracy.
    The blue point plots are derived from reports in SparseOcc \cite{liu2023sparseocc}, while the yellow ones are obtained through our own testing with same setting, i.e., the frames per second (FPS) were conducted using a A100 with PyTorch FP32 backend.
    }
    \label{fig:fig1}
    \bigskip}                   
\makeatother

\maketitle
\input{sec/0_abstract}

\input{sec/1_intro}
\input{sec/2_related}

\input{sec/3_method}

\input{sec/4_experiment}

\input{sec/5_conclusion}
\newpage
{
    \small
    \bibliographystyle{ieeenat_fullname}
    \bibliography{main}
}


\end{document}

%% file: sec/0_abstract.tex
\begin{abstract}
Panoptic occupancy poses a novel challenge by aiming to integrate instance occupancy and semantic occupancy within a unified framework. 
However, there is still a lack of efficient solutions for panoptic occupancy.
In this paper, we propose Panoptic-FlashOcc, a straightforward yet robust 2D feature framework that enables real-time panoptic occupancy. 
Building upon the lightweight design of FlashOcc, 
our approach simultaneously learns semantic occupancy and class-aware instance clustering in a single network, 
these outputs are jointly incorporated through panoptic occupancy procession for panoptic occupancy.
This approach effectively addresses the drawbacks of high memory and computation requirements associated with three-dimensional voxel-level representations. 
With its straightforward and efficient design that facilitates easy deployment, 
Panoptic-FlashOcc demonstrates remarkable achievements in panoptic occupancy prediction. 
On the Occ3D-nuScenes benchmark, it achieves exceptional performance, with 38.5 RayIoU and 29.1 mIoU for semantic occupancy,
operating at a rapid speed of 43.9 FPS. 
Furthermore, it attains a notable score of 16.0 RayPQ for panoptic occupancy, accompanied by a fast inference speed of 30.2 FPS.
These results surpass the performance of existing methodologies in terms of both speed and accuracy.
The source code and trained models can be found at the following github repository: \url{https://github.com/Yzichen/FlashOCC}.

\end{abstract}

%% file: sec/1_intro.tex
\section{Introduction}
\label{sec:intro}
Panoptic occupancy in multi-view sense plays a critical role in autonomous robot navigation \cite{huang2023voxposer}, environment mapping \cite{wang2023embodiedscan}, and self-driving systems \cite{yu2023flashocc,wei2023surroundocc,cao2022monoscene,wang2023panoocc}. It has recently garnered significant attention due to the introduction of the panoptic quality metric \cite{liu2023sparseocc}. 
Panoptic occupancy divides 3D scenes into structured voxels from visual images, and each voxel is assigned with an instance ID,
where each voxel in ``thing" classes is identified by both class labels and instance IDs, 
while the voxel in ``stuff" classes is labeled with only class labels.

Mutil-view 3D panoptic occupancy is still a nascent field and remains an open research problem. 
As its heavy calculation, to date, only one paper, SparseOcc \cite{liu2023sparseocc}, has addressed this problem in a sparse manner. 
3D panoptic occupancy presents challenges in network design as it classifies the segmentation of 3D voxels into semantically segmented regions, as well as the distinction of individual instances. 
Moreover, panoptic occupancy finds its primary application in fields as mentioned at the begin of the article, all of which demand real-time inference and high accuracy while being deployable on various edge chips. 
These challenges have motivated us to seek a more suitable architecture that can address these issues and achieve fast inference speed without compromising accuracy.

Inspired by the success of the bottom-up paradigm in 2D image panoptic segmentation, where semantic segmentation predictions are typically obtained before grouping ``thing" pixels into clusters to identify instances \cite{cheng2020panoptic,yang2019deeperlab,gao2019ssap,uhrig2018box2pix,kirillov2019panoptic}, 
we aim to develop a simple, effective, and deployable network for panoptic occupancy. 
Our proposed approach integrates semantic occupancy with class-aware instance clustering to process panoptic occupancy. 
To ensure inference speed without sacrificing accuracy, 
we adopt the architecture of FlashOcc \cite{yu2023flashocc} to estimate semantic occupancy. 
FlashOcc utilizes a channel-to-height transformation to effectively convert the flattened bird's-eye view (BEV) feature into 3D occupancy predictions, 
without the need for computationally expensive 3D voxel-level representations. 
We then incorporate a lightweight centerness head, 
inspired by Panoptic-DeepLab \cite{cheng2020panoptic}, 
to generate class-aware instance centers. 
The predictions from the semantic occupancy estimation and the centerness head are fused through the panoptic occupancy processing to generate the final panoptic occupancy. 
This results in a highly efficient bottom-up panoptic occupancy network design, which we refer to as Panoptic-FlashOcc.

We evaluated our approach on the challenging Occ3D-nuScenes dataset using three metrics: PayIoU, mIoU, and RayPQ. 
Thanks to its efficient design, Panoptic-FlashOcc achieves state-of-the-art performance whitout whistles and bells, as depicted in Figure \ref{fig:fig1}. 
It achieves the highest performance with 38.5 RayIoU, 31.6 mIoU, and 16.0 RayPQ, while maintaining an inference speed of 35.6, 35.6 and 30.2 FPS respectively. Furthermore, in term of RayIoU, it achieves comparable performance to the best competitor on RayIoU while maintaining the fastest inference speed of 43.9 FPS.

%% file: sec/2_related.tex
\section{Related Work}

\textbf{Panoptic segmentation.}
Since the introduction of panoptic segmentation in Kirillov et al. \cite{kirillov2019panoptic}, numerous efforts have emerged in this domain. 
Initially, adaptations to existing networks involved adding either a semantic \cite{kirillov2019panoptic} or an instance branch \cite{cheng2020panoptic} to state-of-the-art models, 
followed by hand-crafted post-processing techniques \cite{liu2019end,xiong2019upsnet,yang2019deeperlab}.
With the integration of transformer into computer vision, 
researchers have started exploring architectures that can address the panoptic segmentation task in a more unified manner. 
MaskFormer \cite{maskformer} utilizes queries to predict object masks and stuff masks in a unified approach. 
Mask2Former \cite{mask2former} introduces masked-attention, which significantly improves performance on smaller objects by masking unrelated parts of the image. 
While transformer-based methods have shown superior performance compared to previous models, 
they present challenges in terms of deployment on various edge chips.
Recently, the efficient MaskConver \cite{rashwan2024maskconver} has surpassed the above transformer-based models by learning instance centers for both ``thing" and ``stuff" classes using only fully convolutional layers.
This motivates us to seek an efficient and deploy-friendly model that operates solely in a buttom-up fully convolutional manner.

\textbf{Efficient Panoptic occupancy.}
Panoptic occupancy represents an emerging direction that remains under-unexplored, in contrast to the flourishing development in panoptic segmentation. 
Sparseocc \cite{liu2023sparseocc} stands as the first and only study that focuses on both enhancing panoptic quality and promoting inference speed. 
It provides reports on quality and latency using an A100 GPU.
Since semantic occupancy is a sub-task of panoptic occupancy, and panoptic understanding can be expanded experientially from semantic occupancy, we also explore semantic occupancy to identify studies with efficient architectures. 
Dense 3D voxel-level representation has been employed in many works \cite{bevdetocc,fbocc,pan2023renderocc,wang2023panoocc,wei2023surroundocc} to perform occupancy computation, 
albeit requiring computational 3D convolutions or transformer modules. 
Consequently, several studies aim to simplify the models to reduce computational time.
TPVFormer \cite{tpvformer} proposes to use a tri-perspective view representations to supplement vertical structural information, 
where the voxel-level representation is simplified.
VoxFormer \cite{li2023voxformer} utilizes a sparse-to-dense MAE module to complete the occupancy by the sparse queries projected from the perspective view.
SparseOcc further optimizes the occupancy prediction in a full sparse manner \cite{liu2023sparseocc}.

However, the aforementioned methods all adopt the paradigm of 3D voxel-level representation to perform occupancy prediction inevitablely, relying on 3D feature or transformer module.
This design poses challenges for deploying them on edge chips, expect for the Nvidia's solutions.
Inspired by the efficiency and deployability of FlashOcc \cite{yu2023flashocc}, which introduces a channel-to-height module to transform flattened BEV features into 3D semantic occupancy predictions using only 2D convolutions, 
we incorporate FlashOcc into our approach, which enables us to explore efficient and chip-independent panoptic occupancy.

\textbf{Center Point Representation.}
Convolution-based tasks such as detection \cite{huang2021bevdet,fcos,centerpoint}, tracking \cite{zhang2022bytetrack}, instance segmentation \cite{cheng2022sparse}, and panoptic segmentation \cite{cheng2020panoptic,kirillov2019panoptic} rely heavily on the notion of center point prediction.
In the context of panoptic segmentation, Panoptic-DeepLab \cite{cheng2020panoptic} predicts the mass center to represent each instance. Similarly, MaskConver \cite{rashwan2024maskconver} utilizes the center point to model the mask centers for both ``thing" and ``stuff" classes, rather than focusing solely on box centers.

Building upon the above approaches, we extend the representation to include class-aware instance center prediction for each ``thing" class. Experimental results demonstrate that this simple yet efficient representation achieves state-of-the-art panoptic occupancy results on the Occ3D-nuScenes dataset.

%% file: sec/3_method.tex
\begin{figure*}
\centering
	\includegraphics[width=0.9\linewidth]{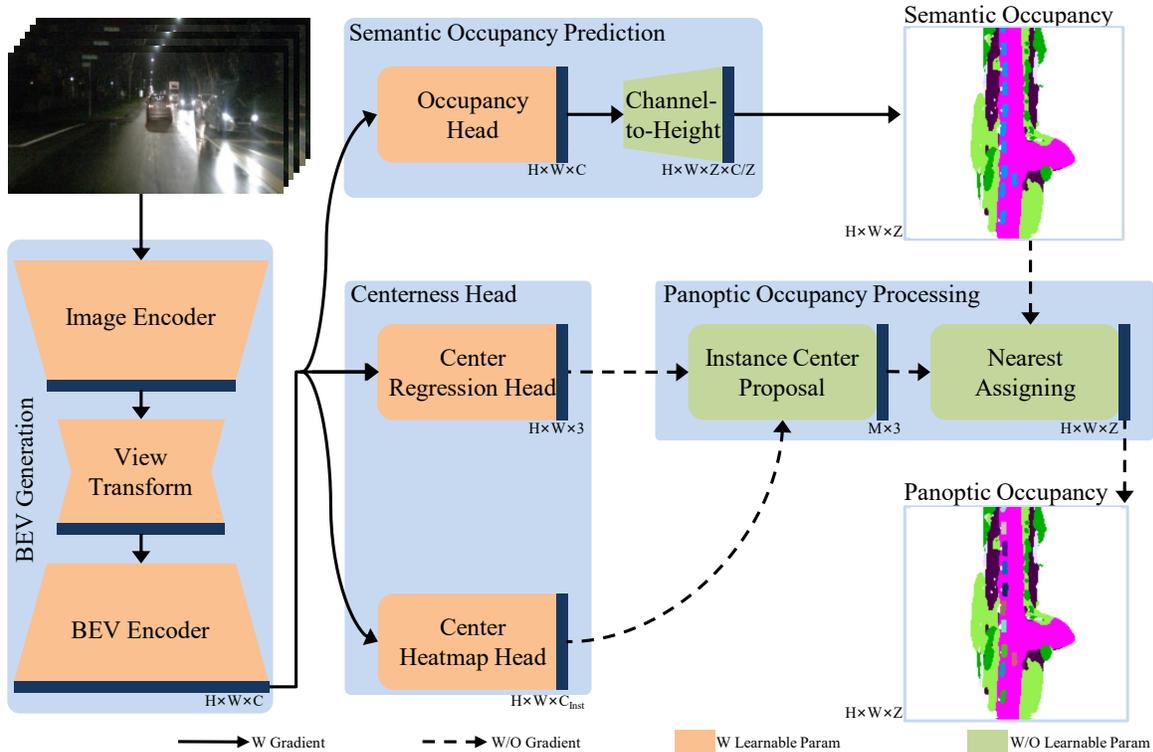}
	\caption{
 The overall architecture of our Panoptic-FlashOcc. 
 The BEV generation and semantic occupancy prediction directly inherit the model structure from FlashOcc, while the newly added centerness head and panoptic occupancy processing enhances semantic occupancy prediction to panoramic occupancy prediction.
 Best viewed in color and zoom.
	}
	\label{fig:arch}
\end{figure*}

\section{Architecture}
In this section, we outline how the proposed instance centers are utilized to integrate the panoptic property into the semantic occupancy task. 
We begin by providing an overview of the architecture in Section \ref{sec:over-arch}. 
Then, we delve into the occupancy head, which predicts the segmentation label for each voxel, in Section \ref{sec:sem-occ-head}. Subsequently, we elaborate on the centerness head, which is employed to generate class-aware instance centers, in Section \ref{sec:centerness-head}. Finally, we describe the panoptic occupancy processing, which serves as an efficient post-processing module for generating panoptic occupancy, in Section \ref{sec:panoptic-occupancy-processing}.

\subsection{Overview Architecture}
\label{sec:over-arch}
As depicted in Fig. \ref{fig:arch}, Panoptic-FlashOcc comprised four main components: BEV generation, semantic occupancy prediction, centerness head and panoptic occupancy processing.
The BEV generation module transforms the surround-view images into the BEV feature $\boldsymbol{F_{BEV}} \in \mathbb{R}^{H \times W \times C}$, where $H$, $W$ and $C$ denote the height, width and channel dimensions of the feature, respectively. This transformation is achieved by employing an image encoder, view transform and BEV encoder, which can be adopted directly from \cite{yu2023flashocc,bevformer,fbocc,pan2023renderocc}.
To ensure efficient deployment on edge chips, we adhere to the configuration of FlashOCC \cite{yu2023flashocc}, where ResNet50 \cite{resnet} is employed as the image encoder, LSS \cite{huang2021bevdet,lss} serves as the view transformer, ResNet18 and FPN are utilized as the BEV encoder.

The semantic occupancy prediction module takes the flatten BEV feature $\boldsymbol{F_{BEV}}$ above as input and generates the semantic occupancy result $\boldsymbol{OCC_{sem}} \in \mathbb{R}^{H \times W \times Z}$, where $Z$ denotes the voxel number perpendicular to the BEV plane.
Meanwhile, the centerness head generates the class-aware heatmap $\boldsymbol{O_{Heat}} \in \mathbb{R}^{H \times W \times C_{Inst}}$ and the regression tensor $\boldsymbol{O_{R}} \in \mathbb{R}^{H \times W \times 3}$ for instance center respectively, where $C_{Inst}$ represents the semantic number of ``thing" class.

Finally, the combination of semantic occupancy result $\boldsymbol{OCC_{sem}}$ and the instance center informations above go through the panoptic occupancy processing for the panoptic prediction $\boldsymbol{OCC_{pano}} \in \mathbb{R}^{H \times W \times Z}$. 
It is important to note that the panoptic occupancy processing serves as a post-processing and does not involve any gradient backpropagation.

\subsection{Semantic Occupancy Prediction}
\label{sec:sem-occ-head}
To ensure a lightweight and easily deployable solution, the architecture of the semantic occupancy prediction module is directly inherited from FlashOCC \cite{yu2023flashocc}. It consists of an occupancy head and a channel-to-height operator, enabling the prediction of semantic labels for both ``thing" and ``stuff" classes.
The occupancy head is a submodule that comprises three 2D convolutional layers. 
In line with the loss setting proposed in \cite{fbocc, wang2023openoccupancy}, the loss function improves upon the pixel-wise cross-entropy loss utilized in FlashOcc by incorporating a distance-aware focal loss ($L_{fl}$) \cite{lin2017focal}. Additionally, to enhance the capacity of 3D Semantic Scene Completion (SSC) in addressing occluded regions, a semantic affinity loss ($L_{scal}^{sem}$) \cite{cao2022monoscene} and the geometry affinity loss ($L_{scal}^{geo}$) are adopted. 
Furthermore, the lovasz-softmax loss ($L_{ls}$) \cite{wang2023openoccupancy} is also introduced into the training framework.

\subsection{Centerness Head}
\label{sec:centerness-head}
The centerness head, proposed within our framework, serves two fundamental purposes:
1) Enhancing the distinction between various objects. By generating sharper semantic boundaries, the centerness head improves the performance of the semantic occupancy branch as an auxiliary submodule. This enhancement is achieved without requiring any additional computational resources during inference when performing only semantic occupancy prediction. This empirical verification is discussed in the subsequent section.
2) Proposing the class label and 3D location of instance center for panoptic occupancy processing. 
In urban scenes, ``thing" objects are generally separated from each other in the flattened bird's eye view (BEV) perception, so the ``thing" centers generated from the BEV features identify with that from the 3D voxel features. 
As illustrated in the bottom center block of Fig. \ref{fig:arch}, 
the centerness head comprises center regression head and center heatmap head. Both modules comprise three convolution layers paired with a $3 \times 3$ kernel.

\paragraph{Center Heatmap Head.}
The significance of center point representation for both ``things" and ``stuff" has been extensively demonstrated in various studies, including object detection \cite{fcos, centerpoint, huang2021bevdet, pointpillar}, instance segmentation \cite{cheng2022sparse}, and panoptic segmentation \cite{rashwan2024maskconver, cheng2020panoptic}.
During the training process, the ground truth instance centerness values are encoded using a 2D gaussian distribution with a standard deviation equal to the diagonal size of the instance mass. The focal loss is employed to minimize the discrepancy between the predicted class-aware heatmap $\boldsymbol{O_{Heat}}$ and the corresponding ground truth heatmap.

\paragraph{Center Regression Head.}
The objective of this module is to estimate the 3D regression tensor $(x, y, z)$ from each pixel to its corresponding instance center of mass. 
Given that the BEV feature is a 2D flattened pillar representation rather than a 3D voxel representation, 
the regression for $x$ and $y$ represents the transitional ratio from the actual center to the discrete index, 
while the regression for $z$ accounts for the estimation of the absolute height ratio.
During the training process, we employ the smooth L1 loss for regression prediction, which is activated only at pixels that belong to object instances.

\subsection{Panoptic Occupancy Processing}
\label{sec:panoptic-occupancy-processing}
The panoptic occupancy processing module serves as an instance label assignment module, designed to be both simple and effective. It exclusively relies on matrix operations and logical operations, without the inclusion of any trainable parameters. This design allows for a straightforward and efficient implementation of the panoptic occupancy processing.

\paragraph{Instance Center Proposal.}

Given the class-aware heatmap $\boldsymbol{O_{Heat}}$, we extract the candidate instance center indexs with local maximum confidence. This is achieved by applying the maxpool operator with a kernel size of $3 \times 3$ to $\boldsymbol{O_{Heat}}$, 
and retaining the indexs whose values remain unchanged before and after maxpooling. 
This process is analogous to non-maximum suppression (NMS) in object detection. 
Subsequently, the indexs with the top 100 confidence scores are retained, and a sequential score threshold $\tau$ (set to 0.3) is utilized to filter out indexs with low confidence.
Finally, we obtain $M$ instance center index proposals $\{(i_{n},j_{n},k_{n}), {n \in {1,...,M}}\}$, where $i$, $j$, and $k$ represent the indexs along the $x$, $y$, and $z$ axis, respectively. The value of $k_{n}$ corresponds to the semantic label of the corresponding instance.

Using the center regression tensor $\boldsymbol{O_{R}}$, we can further obtain the instance center proposals paired with precise 3D locations and semantic labels, denoted as $\Phi=\{{(x_{n},y_{n},z_{n},k_{n}), n \in {1,...,M}\}}$:
\begin{equation}\label{eq1}
\left\{
            \begin{array}{lr}
            x_{n}=\delta x \times (i_{n}+O_{R}[i_{n},j_{n},0])\\
            y_{n}=\delta y \times (j_{n}+O_{R}[i_{n},j_{n},1])\\
            z_{n}=R_{Z} \times O_{R}[i_{n},j_{n},2]).
             \end{array}
\right.
\end{equation}
where $\delta x$ and $\delta y$ are the voxel size along $x$ and $y$ axis, $R_{Z}$ is the perception range along the z axis.

\paragraph{Nearest Assign.} 
We employ a simple nearest assignment module to determine the instance ID for each voxel in $OCC_{sem}$. 
The pseudo-code outlining the processing steps is presented in Algorithm~\ref{algo:alogtithm}. 
Given the semantic occupancy $\boldsymbol{OCC_{sem}}$ and instance centers $\Phi$ as inputs, 
the nearest assignment module outputs the panoptic occupancy $\boldsymbol{OCC_{pano}}$.
First, we initialize the instance ID number $N_{ID}$ to 0. 
For each class $d$ in the semantic labels (with a total of $N_S$ semantic classes), 
we begin by collecting the index set $I_d$ whose value in $\boldsymbol{OCC_{sem}}$ is $d$. 
We then proceed differently depending on whether $d$ belongs to a ``stuff" object or a ``thing" object.

In the case of a ``stuff" object, we assign the value $N_{ID}$ to all indexs in $\boldsymbol{OCC_{pano}}$ corresponding to the index set $I_d$. Afterwards, we increment $N_{ID}$ by 1, as a ``stuff" object is treated as a single instance. (Line 5-8 in Algorithm~\ref{algo:alogtithm})

For ``thing" objects, we first filter out $L$ instance centers $\widetilde{\phi}$ from $\Phi$ whose instance class is $d$. 
For each index $(i,j,k)$ in $I_d$, we select the nearest instance center with the same label from $\widetilde{\phi}$ using the following equation:
\begin{equation}\label{eq2}
    l=\operatorname*{argmin}_l ||\widetilde{\phi}_{l}[:3]-(\delta x \times i,\delta y \times j,\delta z \times k)||^2
\end{equation}
where $\delta z$ is the voxel size along $z$ axis and $l$ is the select index from $\widetilde{\phi}$. Then the value in $\boldsymbol{OCC_{pano}}$ located by index $(i,j,k)$ is set to be $N_{ID}+l$. 
After iterating through all indexs in set $I_d$, we update the instance $N_{ID}$ by adding $L$ to itself. (Line 11-17 in Algorithm~\ref{algo:alogtithm})

Once all semantic labels have been processed, we obtain the final panoptic occupancy prediction $\boldsymbol{OCC_{pano}}$.
In contrast to previous studies for panoptic segmentation that utilize class-agnostic instance centers, the computational complexity of Eq. \ref{eq2} is $O(NWH)$, where $N$ is the number of instance objects across all labels. 
When extending the task from 2D segmentation to voxel-level occupancy, the complexity increases to $O(NWHZ)$. 
However, by utilizing class-aware instance centers, the complexity is reduced to $O(N_{s}WHZ)$, where $N_s$ is the number of objects with a single class label.

\begin{algorithm}[!h]
\SetAlgoLined
\DontPrintSemicolon
\SetNoFillComment
\footnotesize
\KwIn{Semantic occupancy $\boldsymbol{OCC_{sem}}$; instance center $\Phi$}
\KwOut{Panoptic occupancy $\boldsymbol{OCC_{pano}}$}
Initialization: $\boldsymbol{OCC_{pano}} \leftarrow \boldsymbol{0}_{H \times W \times Z}$; $N_{ID} \leftarrow 0$ \;
\For{$d$ in $1,2,...,N_S$}{
    $I_{d} \leftarrow$ the index whose value in $\boldsymbol{OCC_{sem}}$ is $d$ \;
	\If{$d \in stuff$}{
        \For{$(i,j,k)$ in $I_{d}$}{
            $\boldsymbol{OCC_{pano}}[i,j,k] \leftarrow  N_{ID}$;
        }
        $N_{ID} \leftarrow N_{ID}+1$ \;
	}
	\Else{
        $\widetilde{\phi} \leftarrow$ the instance center from $\Phi$ whose semantic class is $d$ \;
        $L \leftarrow$ total instance number in $\widetilde{\phi}$ \;
        \For{$(i,j,k)$ in $I_{d}$}{
            select the nearest instance center indexed $l$ from $\widetilde{\phi}$ with same label by Eq. \ref{eq2} \;
            $\boldsymbol{OCC_{pano}}[i,j,k] \leftarrow  N_{ID} + l$;
        }
        $N_{ID} \leftarrow N_{ID}+L$ \;
    }
}
Return: $\boldsymbol{OCC_{pano}}$
\caption{Pseudo-code of Nearest Assign.}
\label{algo:alogtithm}
\end{algorithm}

%% file: sec/4_experiment.tex
\section{Experiment}

\begin{table*}[t]
  \setlength{\tabcolsep}{0.0145\linewidth}
  \footnotesize
  \centering
  \caption{The performance of 3D semantic occupancy on Occ3D-nuScenes \cite{tian2023occ3d}. 
  }
  \vspace{-0.3cm}
  \label{table:occ3d-nus}
  \scalebox{0.92}{
  \begin{tabular}{l|ccc|c|c|ccc|c|c}
    \toprule[1.5pt]
    Method & Backbone & Input Size & Epochs & Vis. Mask & {RayIoU} & RayIoU\textsubscript{1m} & RayIoU\textsubscript{2m} & RayIoU\textsubscript{4m} & mIoU & FPS \\
    \midrule
    MonoScene~\cite{cao2022monoscene}   & R101$\ast$ & 928$\times$1600 & 24 & $\checkmark$ & - & - & - & - &  6.0 & - \\
    OccFormer~\cite{zhang2023occformer} & R101$\ast$ & 928$\times$1600 & 24 & $\checkmark$ & - & - & - & - & 21.9 & - \\
    TPVFormer~\cite{huang2023tri}       & R101$\ast$ & 928$\times$1600 & 24 & $\checkmark$ & - & - & - & - & 27.8 & - \\
    CTF-Occ~\cite{tian2023occ3d}        & R101$\ast$ & 928$\times$1600 & 24 & $\checkmark$ & - & - & - & - & 28.5 & - \\
    RenderOcc~\cite{pan2023renderocc}   & SwinB      & 512$\times$1408 & 24 & $\checkmark$ & - & - & - & - & 26.1 & - \\
    BEVFormer \cite{bevformer}          & R101       & 1600 $\times$ 900 & 24 & $\checkmark$ & 32.4 & 26.1 & 32.9 & 38.0 & 39.3 & 3.0 \\
    FB-Occ (16f) \cite{fbocc}           & R50        & 704 $\times$ 256  & 90 & $\checkmark$ & 33.5 & 26.7 & 34.1 & 39.7 & 39.1 & 10.3 \\
    FlashOcc \cite{yu2023flashocc}      & R50       & 704 $\times$ 256 & 24 & $\checkmark$ & - & - & - & - & 32.0 & 29.6 \\
    \midrule
    BEVFormer \cite{bevformer}          & R101       & 1600 $\times$ 900 & 24 &      -       & 33.7 &  -   &  -   &  -   & 23.7 & 3.0 \\
    FB-Occ (16f) \cite{fbocc}           & R50        & 704 $\times$ 256  & 90 &      -       & 35.6 &  -   &  -   &  -   & 27.9 & 10.3 \\
    RenderOcc \cite{pan2023renderocc}   & Swin-B     & 1408 $\times$ 512 & 12 &      -       & 19.5 & 13.4 & 19.6 & 25.5 & - & - \\
    BEVDetOcc \cite{bevdetocc}          & R50        & 704 $\times$ 256  & 90 &      -       & 29.6 & 23.6 & 30.0 & 35.1 & - & 2.6\\
    BEVDetOcc (8f)                      & R50        & 704 $\times$ 384  & 90 &      -       & 32.6 & 26.6 & 33.1 & 38.2 & - & 0.8\\
    SparseOcc (8f) \cite{liu2023sparseocc} & R50    & 704 $\times$ 256  & 24 & - & 34.0 & 28.0 & 34.7 & 39.4 & 30.6 & 17.3 \\
    SparseOcc (16f) \cite{liu2023sparseocc}& R50    & 704 $\times$ 256  & 24 & - & 35.1 & 29.1 & 35.8 & 40.3 & 30.9 & 12.5 \\
    \midrule
    Panoptic-FlashOcc-tiny (1f)      & R50    & 704 $\times$ 256  & 24 & - & 34.8 & 29.1 & 35.7 & 39.7 & 29.1 & 43.9 \\
    Panoptic-FlashOcc (1f)           & R50    & 704 $\times$ 256  & 24 & - & 35.2 & 29.4 & 36.0 & 40.1 & 29.4 & 38.7 \\
    Panoptic-FlashOcc (2f)           & R50    & 704 $\times$ 256  & 24 & - & 36.8 & 31.2 & 37.6 & 41.5 & 30.3 & 35.9 \\
    Panoptic-FlashOcc (8f)           & R50    & 704 $\times$ 256  & 24 & - & 38.5 & 32.8 & 39.3 & 43.4 & 31.6 & 35.6 \\
    
    \bottomrule[1.5pt]
  \end{tabular}
}
\label{tab:sota_on_sem}
\end{table*}

\begin{table*}[t]
  \setlength{\tabcolsep}{0.019\linewidth}
  \centering
  \footnotesize
  \caption{The performance of 3D panoptic occupancy on Occ3D-nuScenes \cite{tian2023occ3d}. 
  }
  \vspace{-0.3cm}
  \label{table:occ3d-nus}
  \scalebox{0.92}{
  \begin{tabular}{l|ccc|c|c|ccc|c}
    \toprule[1.5pt]
    Method & Backbone & Input Size & Epochs & Vis. Mask & {RayPQ} & RayPQ\textsubscript{1m} & RayPQ\textsubscript{2m} & RayPQ\textsubscript{4m} & FPS \\
    \midrule
    SparseOcc (8f) \cite{liu2023sparseocc} & R50    & 704 $\times$ 256  & 24 & - & 14.1 & 10.2 & 14.5 & 17.6 & 17.3 \\
    \midrule
    Panoptic-FlashOcc-tiny (1f)      & R50    & 704 $\times$ 256  & 24 & - & 12.9 & 8.8 & 13.4 & 16.5 & 39.8 \\
    Panoptic-FlashOcc (1f)           & R50    & 704 $\times$ 256  & 24 & - & 13.2 & 9.2 & 13.5 & 16.8 & 35.2 \\
    Panoptic-FlashOcc (2f)           & R50    & 704 $\times$ 256  & 24 & - & 14.5 & 10.6 & 15.0 & 18.0 & 30.4 \\
    Panoptic-FlashOcc (8f)           & R50    & 704 $\times$ 256  & 24 & - & 16.0 & 11.9 & 16.3 & 19.7 & 30.2 \\
    \bottomrule[1.5pt]
  \end{tabular}
}
\label{tab:sota_on_pano}
\end{table*}

Within this section, we begin by providing comprehensive information concerning the benchmark and metrics utilized, along with the training details for our Panoptic-FlashOcc, as documented in Section~\ref{sec:exp_setup}.
Subsequently, Section~\ref{sec:sota_exp} showcases the primary performance of our method compared to other state-of-the-art methods on occupancy prediction.
Following that, in Section~\ref{sec:exp_ablation}, we carry out thorough ablation study to explore the efficacy of each component in our proposed method.

\subsection{Experimental Setup}\label{sec:exp_setup}
\textbf{Benchmark.}
We conducted occupancy on the Occ3D-nuScenes~\cite{tian2023occ3d} datasets.
The data collection vehicle is equipped with one LiDAR, five radars, and six cameras, enabling a comprehensive surround view of the vehicle's environment.
It comprises 700 scenes for training and 150 scenes for validation, the duration of each sence is 20s, and groundtruth label is annotated every 0.5s. 
The dataset covers a spatial range of -40m to 40m along the $x$ and $y$ axis, and -1m to 5.4m along the $z$ axis. The occupancy labels are defined using voxels with size of $0.4m \times 0.4m \times 0.4m$ for 17 categories.
Each driving scene contains 20 seconds of annotated perceptual data captured at a frequency of 2 Hz. 

\textbf{Metrics.}
The mean intersection-over-union (mIoU) and RayIoU are utilized to report the semantic segmentation performance,
the later aims to solve the inconsistency penalty along depths raised in
traditional mIoU criteria \cite{liu2023sparseocc}.
and the metric of panoptic quality (PQ) is introduced to evaluate the accuracy of panoptic segmentation.

\begin{figure*}
\centering
    \begin{overpic}[width=1.0\textwidth]{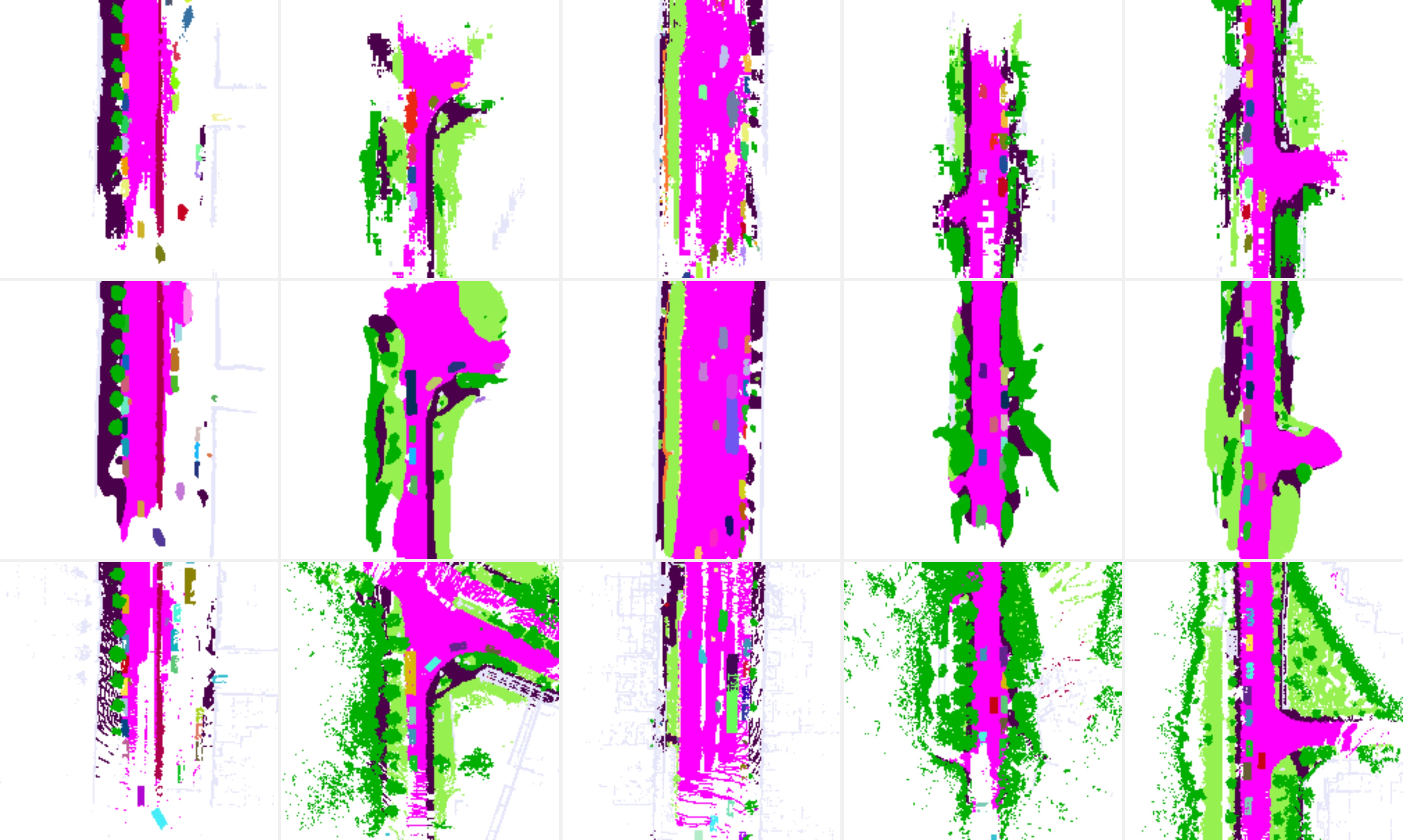}
        \put(0, 47){\rotatebox{90}{\color{black}{\footnotesize SparseOcc}}}
        \put(0, 29){\rotatebox{90}{\color{black}{\footnotesize Ours}}}
        \put(0, 5){\rotatebox{90}{\color{black}{\footnotesize Ground Truth}}}
    \end{overpic}
	\caption{
 Qualitative results of panoptic occupancy on Occ3D-nuScenes.
	}
	\label{fig:comparsion_on_pano}
\end{figure*}

\begin{figure*}
\centering
    \begin{overpic}[width=1.0\textwidth]{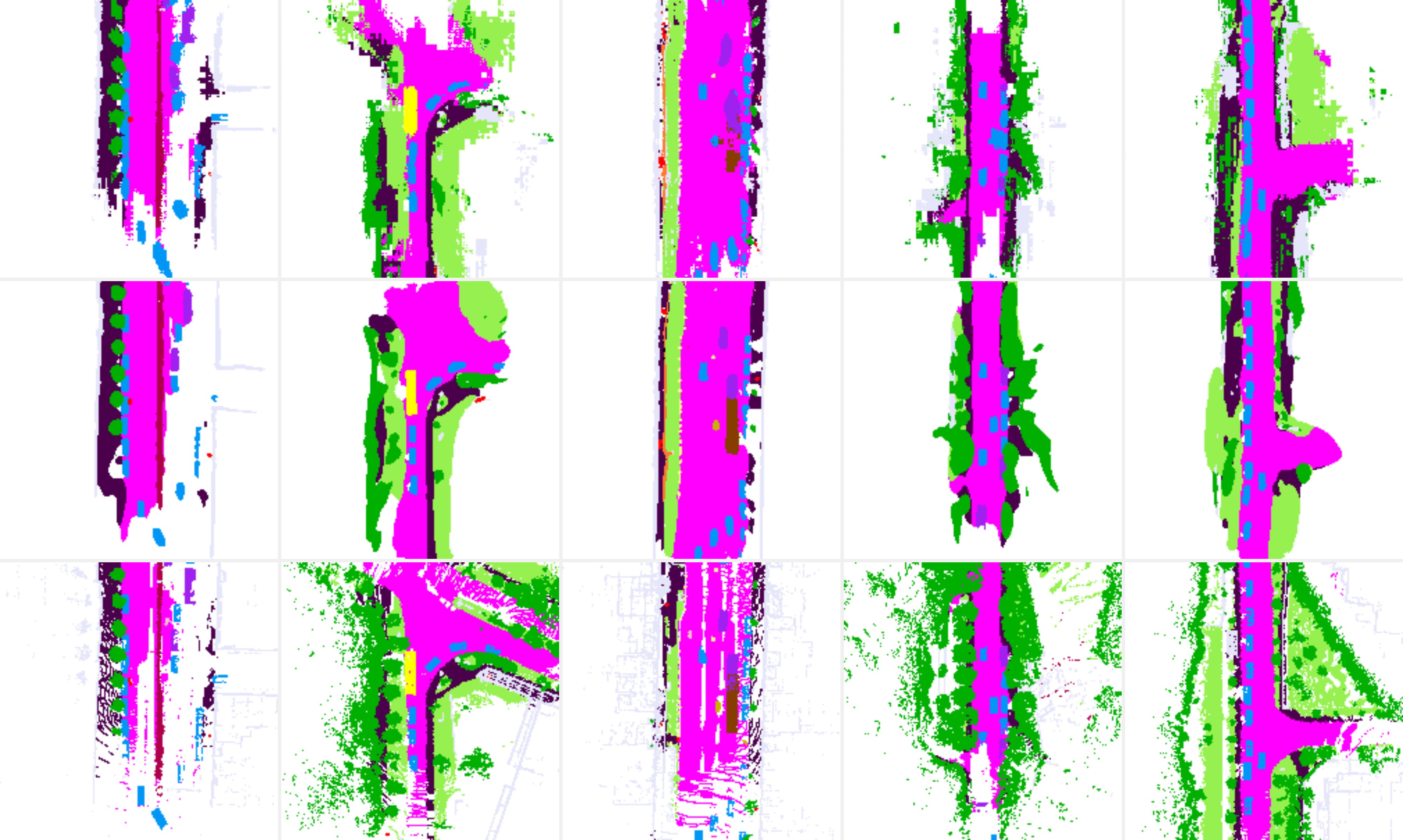}
        \put(0, 47){\rotatebox{90}{\color{black}{\footnotesize SparseOcc}}}
        \put(0, 29){\rotatebox{90}{\color{black}{\footnotesize Ours}}}
        \put(0, 5){\rotatebox{90}{\color{black}{\footnotesize Ground Truth}}}
    \end{overpic}
	\caption{
 Qualitative results of semantic occupancy on Occ3D-nuScenes.
	}
	\label{fig:comparsion_on_sem}
\end{figure*}

\textbf{Training Details.}
In our paper, we adopt Resnet50 as backbone and image with size of $704 \times 256$ as input.
All our settings are trained using PyTorch with a total batch size of 16 distributed across 4 GPUs, and the total training epoch is set to 24. 
AdamW~\cite{loshchilov2017decoupled} optimizer is adopt with weight decay set to be $1 \times 10^{-2}$,
the learning rate keeps constant as $1 \times 10^{-4}$ except the linear warmup in first 200 iterations,
the weights of each loss in our method are set to be 1.0.
The use of camera mask is eliminated during training. FPS is measured on NVIDIA A100 GPU (PyTorch fp32 backend).
We also provide a "tiny" model variant where the channel numbers in occupancy head are reducing half.

\subsection{Comparison with State-of-the-art Methods}\label{sec:sota_exp}

We conducted a comprehensive comparison between our proposed Panoptic-FlashOcc and other state-of-the-art methods on task of semantic occupancy as well as panoptic occupancy. 

The results for semantic occupancy are presented in Table \ref{tab:sota_on_sem}, where we report the metrics of RayIoU and mIoU along with the corresponding FPS.
In terms of challenging RayIoU, Panoptic-FlashOcc-tiny (1f) outperforms all other methods except SparseOcc (16f), achieving an impressive 43.9 FPS. 
Additionally, Panoptic-FlashOcc (1f) surpasses all previous competitors with a remarkable 38.7 FPS. 
When expanding the input to 2 frames, we observe notable performance improvements, with a 1.7 increase in RayIoU compared to the previous best competitor.
Furthermore, by utilizing 8 frames, we achieve the best performance of 38.5 RayIoU and 31.6 mIoU while maintaining a fast inference speed of 35.6 FPS. 
This outperforms SparseOcc (16f) by 3.4 RayIoU, 0.7 mIoU and 23.1 FPS.

Table \ref{tab:sota_on_pano} presents a comparative analysis between Panoptic-FlashOcc and SparseOcc in panoptic occupancy prediction task. Panoptic-FlashOcc (2f) exhibits an advantage over SparseOcc (8f) with the improvement of 0.4 RayPQ, while maintaining a noteworthy gain of 13.1 FPS in speed.
Furthermore, when the input frames are expanded to same number, Panoptic-FlashOcc (8f) outperforms SparseOcc (8f) by a notable margin of 1.9 RayPQ, while maintaining a faster inference speed of 30.2 FPS.


The qualitative visualization of our method, SparseOcc, and the ground truth for panoptic and semantic occupancy is illustrated in Figures \ref{fig:comparsion_on_pano} and \ref{fig:comparsion_on_sem} respectively. Our method exhibits sharper results in the case of instance objects (such as car, bus, truck, trailer, etc.) compared to SparseOcc.
In fact, our method even outperforms the ground truth where there are numerous gaps in the mass of vehicle and visible driveable surface.


\subsection{Ablation Study}\label{sec:exp_ablation}
We conduct ablative experiments to demonstrate the efficacy of each component in our method. 
Unless otherwise stated, all of the experiment are performed on Panoptic-FlashOcc-tiny (1f).

\textbf{Effect of Losses for Semantic Occupancy.}\label{}
In our semantic occupancy prediction module, distance-aware focal loss $L_{fl}$, semantic affinity loss $L_{scal}^{sem}$,
geometry affinity loss $L_{scal}^{geo}$ and lovasz-softmax loss $L_{ls}$ 
\cite{wang2023openoccupancy} are adopted as the training objective to minimise the discrepancy between prediction and ground truth. 
We aim to determine the crucial contributions of each component to model's performance.
Table \ref{tab:lass_effect} presents the results.
The baseline model with $L_{fl}$ achieves 18.39 RayIou and 15.39 mIoU.
By incorporating $L_{scal}^{sem}$, the performance sinificantly improves to 34.32 RayIou and 28.82 mIoU.
Furthermore, by adding $L_{scal}^{geo}$, the performance further improves to 34.78 RayIou and 29.12 mIoU. 
However, the inclusion of $L_{ls}$ only leads to a marginal improvement, with a gain of 0.04 RayIoU and 0.02 mIoU.
These findings highlight the significant contribution of semantic scene completion, i.e. $L_{scal}^{sem}$ and $L_{scal}^{geo}$, to the prediction of semantic occupancy.

\begin{table}[htb] 
    \footnotesize
    \setlength{\tabcolsep}{3.8mm}
    \centering
    \begin{tabular}{cccc|c|c}
    \toprule[1.5pt]
     $L_{fl}$ & $L_{scal}^{sem}$ & $L_{scal}^{geo}$ & $L_{ls}$ & RayIou & mIoU \\
     \midrule
     \checkmark &  &  &  & 18.39 & 15.39 \\
     \checkmark & \checkmark &  &  & 34.32 & 28.82 \\
     \checkmark & \checkmark & \checkmark &  & 34.78 & 29.12 \\
     \checkmark & \checkmark & \checkmark & \checkmark & 34.82 & 29.14 \\
    \bottomrule[1.5pt]
    \end{tabular}
    \vspace{-0.3cm}
    \caption{The impact of losses in semantic occupancy prediction module. 
    }
    \label{tab:lass_effect}
\end{table}

\textbf{Comparison with Different Instance Scores Threshold.}\label{}
As the panoptic occupancy processing module serves as a post-processing step, we conducted an ablation study to evaluate the impact of the parameter $\tau$. The results are presented in Table \ref{tab:tau_study}.
We began by setting $\tau$ to 0.1 and observed that the best performance was achieved when $\tau$ was set to 0.3. Increasing $\tau$ beyond this value did not lead to any further improvement in performance, as leak detection of instance center increases. 
Therefore, we selected 0.3 as the default setting for $\tau$ in our paper.

\begin{table}[htb] 
    \footnotesize
    \setlength{\tabcolsep}{1.5mm}
    \centering
    \begin{tabular}{c|ccccc}
    \toprule[1.5pt]
     \diagbox{Method}{$\tau$} & 0.1 & 0.2 & 0.3 & 0.4 & 0.5 \\
     \midrule
     Panoptic-FlashOcc-tiny (1f) & 12.59 & 12.80 & \textbf{12.89} & 12.62 & 12.08  \\
     Panoptic-FlashOcc (1f)      & 12.91 & 13.13 & \textbf{13.18} & 12.95 & 12.57  \\
     Panoptic-FlashOcc (2f)      & 14.20 & 14.42 & \textbf{14.52} & 14.22 & 13.95  \\
    \bottomrule[1.5pt]
    \end{tabular}
    \vspace{-0.3cm}
    \caption{Effect of score threshold on panoptic occupancy.}
    \label{tab:tau_study}
\end{table}

\textbf{Panoptic Benefits Semantic Occupancy.}\label{}
This study seeks to provide empirical evidence supporting the enhancement of semantic occupancy capacity through the inclusion of a centerness head, which aggregates features encoded in each instance mass. 
To evaluate this proposition, we conducted an experiment, the details of which are presented in Table. \ref{tab:pano_improve_sem}.
For SparseOcc, the inclusion of the panoptic occupancy task results in a slight degradation of the performance of semantic occupancy, reducing it from the baseline's 35.0 RayIoU to 34.5 RayIoU. 

Unlike SparseOcc, our efficient framework demonstrates the ability to further enhance semantic occupancy when incorporating the centerness head for panoptic prediction.
In the bottom 6 rows of Table \ref{tab:pano_improve_sem}, we examine the impact of panoptic occupancy on three different settings of Panoptic-FlashOcc. 
Notably, the performance of semantic occupancy improves consistently when the panoptic occupancy task is introduced. 
The performance enhancement becomes more significant when using a larger occupancy head (from 0.24 to 0.29 RayIoU and 0.31 to 0.48 mIoU) and when incorporating more history frames as input (from 0.29 to 0.77 RayIoU and 0.48 to 0.74 mIoU).
It is worth mentioning that Panoptic-FlashOcc (2f) outperforms SparseOcc (8f) by 1.26 RayIoU and 0.52 RayPQ, indicating that our proposed method is a more suitable framework for occupancy prediction.

\begin{table}[htb] 
    \footnotesize
    \setlength{\tabcolsep}{0.3mm}
    \centering
    \begin{tabular}{ccc|cc|c}
    \toprule[1.5pt]
     Method & Sem. & Pano. & RayIou & mIou & RayPQ \\
     \midrule
     SparseOcc (8f) & \checkmark &              & 35.0$_{\textcolor{white}{+0.00}}$ & - & - \\
     SparseOcc (8f) & \checkmark &  \checkmark  & 34.5$_{\textcolor{blue}{-0.50}}$ & 14.0$_{\textcolor{white}{+0.00}}$ & - \\
     \midrule
     Panoptic-FlashOcc-tiny (1f) & \checkmark &             & 34.57$_{\textcolor{white}{+0.00}}$ & 28.83$_{\textcolor{white}{+0.00}}$ & - \\
     Panoptic-FlashOcc-tiny (1f) & \checkmark &  \checkmark & 34.81$_{\textcolor{red}{+0.24}}$   & 29.14$_{\textcolor{red}{+0.31}}$ & 12.89 \\
     Panoptic-FlashOcc (1f) & \checkmark &             & 34.93$_{\textcolor{white}{+0.00}}$ & 28.91$_{\textcolor{white}{+0.00}}$ & - \\
     Panoptic-FlashOcc (1f) & \checkmark &  \checkmark & 35.22$_{\textcolor{red}{+0.29}}$   & 29.39$_{\textcolor{red}{+0.48}}$ & 13.18 \\
     Panoptic-FlashOcc (2f) & \checkmark &             & 35.99$_{\textcolor{white}{+0.00}}$ & 29.57$_{\textcolor{white}{+0.00}}$ & - \\
     Panoptic-FlashOcc (2f) & \checkmark &  \checkmark & 36.76$_{\textcolor{red}{+0.77}}$   & 30.31$_{\textcolor{red}{+0.74}}$ & 14.52 \\
    \bottomrule[1.5pt]
    \end{tabular}
    \vspace{-0.3cm}
    \caption{
    The impact of panoptic occupancy to semantic occupancy in different paradigm.
    }
    \label{tab:pano_improve_sem}
\end{table}

%% file: sec/5_conclusion.tex
\section{Limitations}
While our method achieves impressive performance in urban scenes and can be easily extended to semantic occupancy in indoor scenes, applying panoptic occupancy to indoor scenes requires additional effort due to the potential overlap of many objects. One possible approach is to directly extend a new convolutional branch with a channel-to-height operator to predict the instance centers along the height dimension. We leave this challenge in indoor scenes to be a focus of our future work.

\section{Conclusion}
This paper introduces Panoptic-FlashOcc, an efficient and deploy-friendly framework for panoptic occupancy prediction. Building upon the established FlashOcc, our framework enhances semantic occupancy to panoptic occupancy by incorporating the centerness head and panoptic occupancy processing. 
Panoptic-FlashOcc achieves the fastest inference speed and the highest performance among all state-of-the-art approaches on the challenging Occ3D-nuScenes benchmark. 
We hope that the proposed Panoptic-FlashOcc can serve as a strong baseline in 3D panoptic occupancy community.